\documentclass[runningheads]{llncs}
\usepackage[T1]{fontenc}
\usepackage{graphicx}
\usepackage{amsmath}
\usepackage{multirow}
\usepackage{marvosym}
\usepackage{hyperref}
\usepackage{svg}
\usepackage{algorithm}
\usepackage{algpseudocode}
\usepackage{amsmath, amssymb}
\usepackage{xcolor}
\hypersetup{
colorlinks=true,
citecolor=blue,   
linkcolor=blue
}
\begin{document}
\title{ProbPlug: A Plugin Uncertainty Network for Reliable Confidence in LLM Binary Classification}
\titlerunning{ProbPlug: Reliable Confidence in LLM Binary Classification}
%
%
\author{
Jianzong Wang\inst{1}\textsuperscript{$\dagger$} \and
Chuhang Liu\inst{1,2}\textsuperscript{$\dagger$} \and
Botao Zhao\inst{1} \and
Zuheng Kang\inst{1} \and
Xulong Zhang\inst{1} \and
Xiaoyang Qu\inst{1} \and
Junqing Peng\inst{1} \and
Zhiewei Ye\inst{3} \and
Yayun He\inst{1}\textsuperscript{(\Letter)}
}
\authorrunning{J. Wang et al.}
%
\institute{
Ping An Technology (Shenzhen) Co., Ltd., Shenzhen, China, \and
Tsinghua Shenzhen International Graduate School, Shenzhen, China, \and
Hubei University of Technology, Hubei, China \\
\email{heyayun0618@163.com}
}

\maketitle              
\footnotetext[1]{$\dagger$ These authors contributed equally to this work.}
\begin{abstract}
Large language models (LLMs) have achieved strong performance across a broad range of classification settings, yet the reliability of their predictions remains a major obstacle to deployment in high-stakes scenarios. Although confidence estimation for LLMs has been widely studied, confidence calibration for LLM-based classification remains underexplored.
We introduce ProbPlug, a lightweight confidence estimation framework for LLM-based binary classification, which predicts whether an output is correct using internal token features extracted from a frozen LLM. ProbPlug employs a self-attention module to aggregate hidden representations and can be integrated into the original inference pipeline without modifying the base model. Experiments across multiple tasks involving both text-based and multimodal large models show that ProbPlug provides more reliable confidence estimates, improves classification performance with negligible additional overhead, and exhibits strong generalization across tasks. These results indicate that ProbPlug serves as a practical solution for confidence estimation in LLM-based classification. Our code is publicly available at \href{https://github.com/ChuhangLiu2002/ProbPlug}{this repository}.

\end{abstract}
\section{Introduction}
\label{sec:intro}
Large language models (LLMs) have achieved promising results across both textual and multimodal classification tasks. \cite{zhang2025pushing}, they can still produce misleading or incorrect predictions. Unlike conventional classifiers, which provide probability scores that can be used to adjust precision-recall trade-offs, LLM-based methods often lack reliable confidence estimates. This limitation hinders their use in reliability-sensitive applications. Therefore, accurately estimating model confidence is crucial for practical deployment.

In recent years, researchers have explored various approaches to solve the confidence estimation problem for LLMs \cite{mahaut2024factual}, including:
\textbf{(1) Verbalization}: enabling LLMs to directly output confidence scores \cite{tian2023just};
\textbf{(2) Logit-based methods}: leveraging the probability distributions of generated tokens \cite{yin2023large}, with recent work such as LogTokU \cite{ma2025estimating} further analyzing token-level logit uncertainty for confidence estimation;
\textbf{(3) Self-consistency}: taking the consistency of results from multiple repeated queries as the confidence metric \cite{wang2023selfconsistency}, as exemplified by SelfCheckGPT \cite{manakul2023selfcheckgpt};
\textbf{(4) Trained probes}: employing probabilities output by pre-trained classifiers as confidence estimates \cite{azaria2023internal};
\textbf{(5) Hybrid methods}: integrating two or more of the aforementioned approaches (e.g., CISC \cite{taubenfeld2025confidence}).
Directly applying these existing methods to LLM-based classification tasks is undoubtedly a simple and feasible approach. However, since these methods are not specifically designed for this scenario, their performance fails to reach an optimal level. Accordingly, this work focuses on confidence estimation for LLM-based binary classification.

Inspired by findings in neuroscience \cite{warren2008does}, we note that during the human brain's decision-making process, signals flow between different brain regions. For instance, when the human ear perceives a question, sensory organs first receive the information; the signal then propagates from lower-level brain regions to higher-order cortices, where comprehension occurs in Wernicke's Area, analysis in the prefrontal cortex, answer generation in Broca's Area, and finally output via the motor cortex. Notably, the resulting answer and its confidence are associated with each step of this process.

Drawing on this insight, we hypothesize that LLM decision-making is closely related to representations across Transformer layers. Accordingly, we propose \textbf{ProbPlug}, a confidence network that takes layer-wise outputs as input and predicts confidence scores. To improve generalization, we focus on binary classification with \texttt{Yes}''/\texttt{No}'' outputs and use only newly generated tokens as input. Experiments show that ProbPlug provides reliable confidence estimation while maintaining strong task performance and cross-task generalization, combining the generalization strength of LLM-based methods with the calibration ability of non-LLM-based approaches.The main contributions of this work are summarized below:
\begin{itemize}
    \item \textbf{We investigate the problem of confidence estimation in LLM-based classification} and highlight that unreliable confidence remains a major obstacle to deploying LLMs in risk-sensitive scenarios.
    \item \textbf{We propose ProbPlug,} a lightweight plug-and-play framework that derives confidence scores from hidden representations of a frozen LLM, enabling estimation of prediction correctness without altering the base model.
    \item \textbf{Extensive experiments on multiple benchmarks} show that ProbPlug provides better-calibrated confidence estimates, improves downstream classification performance with negligible additional inference overhead, and demonstrates promising cross-task generalization ability.
\end{itemize}

\section{Problem Statement}
\label{sec:format}
\begin{figure}[t!] 
  \centering 
  \includegraphics[width=\textwidth]{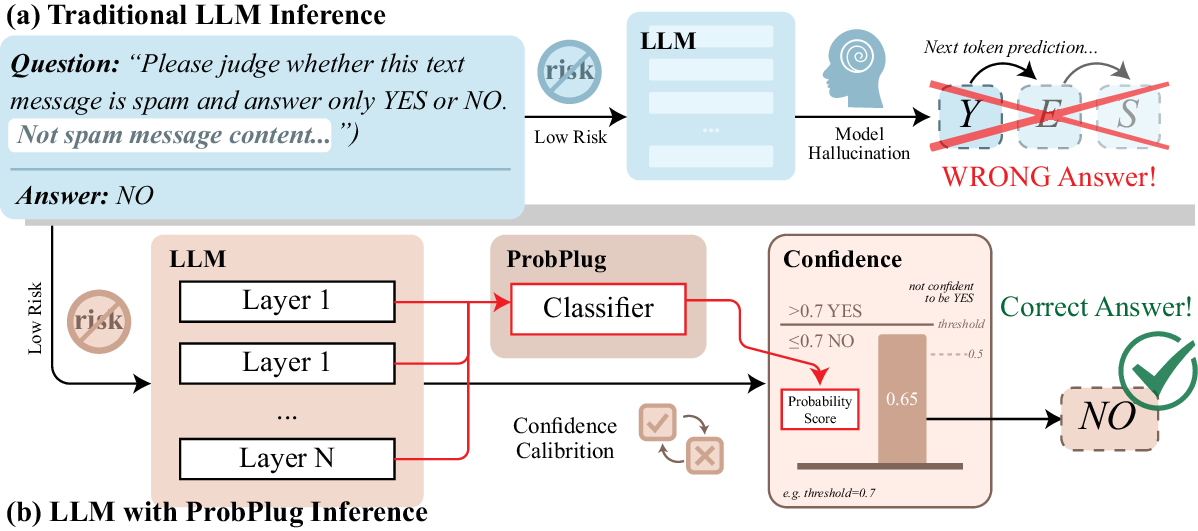} 
  \caption{\textbf{Motivation for the proposed method.} (a) Traditional LLM inference relies on next-token probabilities, which are often poorly calibrated and difficult to threshold in high-stakes settings.
  (b) ProbPlug estimates confidence from multi-layer hidden representations of a frozen LLM, enabling reliable and flexible threshold-based filtering.} 
  \label{fig:motivation}
\end{figure}

As shown in Fig. \ref{fig:motivation}, LLMs can perform classification tasks through prompting, where the model generates a discrete textual response such as ``\texttt{Yes}'' or ``\texttt{No}''. However, such outputs do not naturally provide calibrated confidence scores, limiting their use in reliability-sensitive scenarios.

Given an input prompt $x_{1:t-1}$, the LLM generates the next token $y_{t}$:
\begin{equation}
P(y_{t}|x_{1:t-1}) = \mathrm{LLM}(x_{1:t-1}),
\end{equation}
For binary classification tasks, the generated token is constrained to the set \{\texttt{Yes}, \texttt{No}\}. While it indicates the model's decision, it does not explicitly reflect decision confidence. In this work, we estimate the probability that the generated answer is correct by leveraging hidden representations from different transformer layers, which capture complementary information at different stages of model computation. Let $Z$ denote the hidden representations corresponding to the final generated token across all transformer layers:
\begin{equation}
Z = \{z^{(l)}\}_{l=0}^{L}
\end{equation}
where $L$ denotes the total number of transformer layers. Our goal is to learn a lightweight confidence estimator $f_{\mathrm{prob}}(\cdot)$ such that
\begin{equation}
P(Y = 1| Z) = f_{\text{prob}}(Z)
\end{equation}
where $Y=1$ denotes that the generated answer is correct. Based on this formulation, we propose \textbf{ProbPlug}, a plug-in uncertainty network that estimates decision confidence from layer-wise hidden representations of a frozen LLM. Only a lightweight external network needs to be trained, making ProbPlug plug-and-play and computationally efficient.

\section{Method}
\subsection{Model Architecture}
\label{architecture}
\begin{figure}[t!] 
  \centering 
  \includegraphics[width=0.8\textwidth]{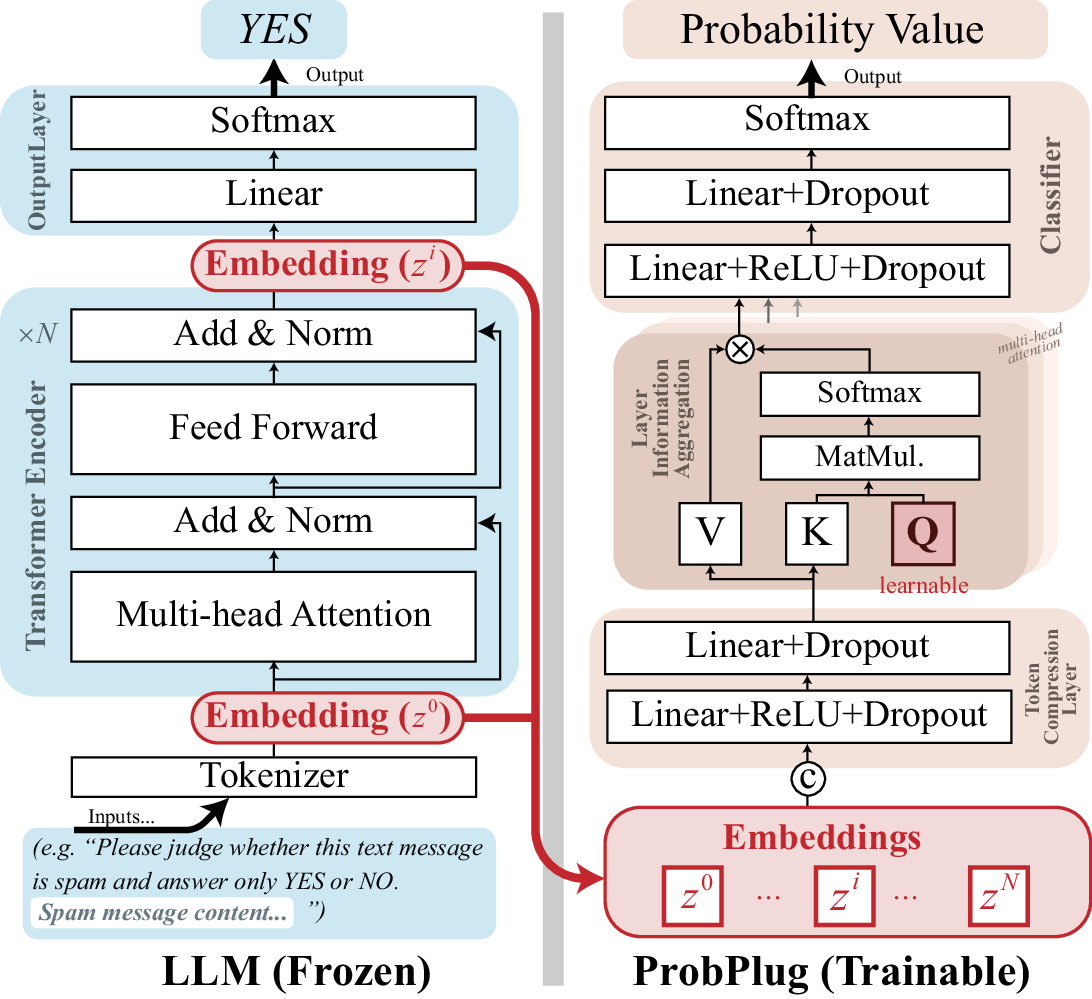} 
  \caption{\textbf{The overview of the proposed method.} The left sub-figure illustrates the frozen large language model, while the right subfigure presents the architecture of our proposed ProbPlug.} 
  \label{fig:overview}
\end{figure}

The hidden representations across Transformer layers encode semantic, reasoning, and decision-related information, which may also reflect model uncertainty. Based on this insight, ProbPlug aggregates these layer-wise signals for confidence estimation. As shown in Fig.~\ref{fig:overview}, the framework is composed of three modules: Token Compression (TC), Layer Information Aggregation (LIA), and a classification head. The backbone LLM remains frozen during training.

\subsubsection{Hidden State Extraction}
Given an input prompt, the frozen LLM produces a binary output token. We extract the hidden representations associated with the final generated token from all transformer layers.
\begin{equation}
Z = [z^{(1)}, z^{(2)}, ..., z^{(L)}]
\end{equation}
Let $Z$ denote the set of hidden states across layers, which has a dimension of $(N, S, L, D)$, where $N$ denotes batch size, $S$ the number of tokens, $L$ the number of distinct layers, and $D$ the feature dimension.
In our default configuration, we select only the final generated token ($S=1$), which directly reflects the model’s decision and empirically provides stronger cross-task generalization ability.

\subsubsection{Token Compression Module}
The extracted hidden representations may contain redundant token-level information. To refine layer-wise features, we introduce a token-wise feature transformation module. For $S=1$, it enhances single-token features via nonlinear mapping.
Subsequently, the token representations are aggregated using two fully connected layers: it first maps the dimension $S$ to $2S$, and then compresses it to $1$.
\begin{equation}
\tilde{z} = W_{2}\sigma(W_{1}z^{(l)}+b_{1})+b_{2}
\end{equation}
where $W_{1}$, $W_{2}$, $b_1$, and $b_2$ denote the trainable parameters. $\sigma$ denotes a nonlinear activation function.
This module compresses token-level features into a single representation per layer, producing $\tilde{Z}$:
\begin{equation}
\tilde{Z} = [\tilde{z^{(1)}}, \tilde{z^{(2)}}, ..., \tilde{z^{(L)}}],
\end{equation}
which summarizes the decision-related information from each transformer layer, has a dimension of $(N, 1, L, D)$.

\subsubsection{Layer Information Aggregation Module}
Different transformer layers contribute unequally to the model’s decision confidence. To adaptively combine these representations, we leverage multi-head attention to adaptively fuse layer-wise representations.
Let
\begin{equation}
Q = W_{Q}q, K = W_{K}\tilde{Z}, V = W_{V}\tilde{Z},
\end{equation}
where $q$ is a learnable query vector.
The attention weights are computed as
\begin{equation}
\alpha = \mathrm{softmax}(\frac{QK^{T}}{\sqrt{d_{k}}})
\end{equation}
The aggregated representation is then obtained as
\begin{equation}
z_{\mathrm{agg}} = \alpha V
\end{equation}
This formulation allows the model to adaptively weight information from different transformer layers during confidence estimation.

\subsubsection{Confidence Prediction}
The aggregated representation $z_{\mathrm{agg}}$ is passed through a lightweight classifier to produce the confidence score:
\begin{equation}
P = \sigma(W_{c}z_{\mathrm{agg}}+b_{c})
\end{equation}
where $W_{c}$ and $b_c$ are learnable parameters. $P$ represents the estimated probability that the LLM’s generated answer is correct.

\subsection{Cross-task Generalization and Multi-class Classification}
A key advantage of this formulation is that it naturally supports cross-task generalization. Since many classification tasks can be reformulated as Yes/No decision problems, the trained 
confidence estimator can be directly applied to unseen tasks by modifying the prompt template.

For multi-class classification tasks, we decompose the task into multiple onv-vs-rest binary decisions. Let $K$ be the total number of classes, and $y_{c}$ denote the confidence score associated with class $c$. The final class probability is computed via a softmax normalization:
\begin{equation}
P(\hat{y}_{c}=c) = \frac{e^{y_{c}}}{\sum_{j=1}^{K} e^{y_{j}}}
\end{equation}
This formulation allows ProbPlug to be extended to multi-class and multimodal classification scenarios.

\begin{table}[t]
\centering
\caption{Performance comparison (Part 1) of our proposed ProbPlug across in-task and cross-task scenarios. Trained on the SMS-SPAM dataset \cite{almeida2011contributions}, the model was evaluated on SST2 \cite{socher2013recursive} and Toxic Comment Classification \cite{borkan2019nuanced}. \textbf{$^*$} indicates the in-task dataset. \textbf{Bold} values denote the best performance.}
\label{TAB1}
\begin{tabular}{|l|c|c|c|c|c|c|}
\hline
\multirow{2}{*}{\bfseries Methods} & \multicolumn{2}{c|}{\bfseries SMS-SPAM$^{*}$} & \multicolumn{2}{c|}{\bfseries SST2} & \multicolumn{2}{c|}{\bfseries Toxic Comment Class} \\
\cline{2-7}
& \bfseries F1-score & \bfseries AUPRC & \bfseries F1-score & \bfseries AUPRC & \bfseries F1-score & \bfseries AUPRC  \\
\hline
Qwen3-8B \cite{yang2025qwen3} & 88.49 & / & 88.75 & / & 81.07 & / \\
Verbalization \cite{tian2023just}  & 87.66 & 82.48 & 90.10 & 86.27 & 81.25 & 63.35 \\
Logit \cite{yin2023large} & 88.64 & 92.02 & 89.44 & 95.28 & 78.91 & 72.42 \\
Self-Consist \cite{wang2023selfconsistency} & 88.44 & 70.13 & 89.55 & 87.18 & 81.66 & 48.63 \\
CISC \cite{taubenfeld2025confidence} & 86.86 & 68.80 & \textbf{91.12} & 93.49 & 81.50 & 50.64 \\
SAPLMA \cite{azaria2023internal} & 92.05 & 95.85 & 87.89 & 96.85 & 81.09 & 70.80 \\
\hline
\textbf{ProbPlug (Ours)} & \textbf{94.78} & \textbf{97.56} & 88.47 & \textbf{97.42} & \textbf{81.68} & \textbf{73.20}  \\
\hline
\end{tabular}
\end{table}

\begin{table}[t]
\centering
\caption{Performance comparison (Part 2) of our proposed ProbPlug across in-task and cross-task scenarios. The model was evaluated on Civil Comments \cite{borkan2019nuanced} and Amazon-polarity \cite{zhang2015character}.}
\label{TAB2}
\begin{tabular}{|l|c|c|c|c|}
\hline
\multirow{2}{*}{\bfseries Methods} & \multicolumn{2}{c|}{\bfseries Civil comments} & \multicolumn{2}{c|}{\bfseries Amazon-polarity} \\
\cline{2-5}
& \bfseries F1-score & \bfseries AUPRC  & \bfseries F1-score & \bfseries AUPRC  \\
\hline
Qwen3-8B \cite{yang2025qwen3} & 66.91 & / & 94.42 & / \\
Verbalization \cite{tian2023just}  & 66.65 & 51.82 & 88.96 & 85.54 \\
Logit \cite{yin2023large} & 65.86 & 55.15 & 94.52 & 97.06  \\
Self-Consist \cite{wang2023selfconsistency} & 66.41 & 45.41 & 94.55 & 94.16 \\
CISC \cite{taubenfeld2025confidence} & 67.09 & 45.51 & 94.10 & 93.70 \\
SAPLMA \cite{azaria2023internal} & 57.19 & 56.50 & 94.46 & 97.65 \\
\hline
\textbf{ProbPlug (Ours)} & \textbf{67.20} & \textbf{58.02} & \textbf{94.72} & \textbf{98.72}  \\
\hline
\end{tabular}
\end{table}

\section{Experimental Results and Analysis}
\label{sec:pagestyle}
\subsection{Experimental Configuration}
\subsubsection{Dataset}
To evaluate generalization, we conduct experiments on five public text datasets: SMS Spam Collection \cite{almeida2011contributions}, SST-2 \cite{socher2013recursive}, Toxic Comment Classification, Civil Comments \cite{borkan2019nuanced}, and Amazon Polarity \cite{zhang2015character}, covering spam detection, sentiment analysis, toxicity detection, and product review classification. We further use IEMOCAP \cite{busso2008iemocap}, which contains 5,531 utterances from four emotion categories, to evaluate performance on multimodal large models.

\subsubsection{Baselines}
We adopt the Qwen3-8B model as the baseline and directly apply it to binary text classification across all datasets. We further explore variants where Qwen outputs either Verbalization \cite{tian2023just} or Logit \cite{yin2023large}. In addition, we compare advanced inference strategies, including Self-Consistency \cite{wang2023selfconsistency} and CISC \cite{taubenfeld2025confidence}. Meanwhile, we evaluate SAPLMA \cite{azaria2023internal}, which attaches and trains a three-layer MLP classification head on top of the model’s internal representations. Model performance is quantified using F1-score and the area under the precision-recall curve (AUPRC), as is common in related literature \cite{kadavath2022language}.

\subsection{Main Performance}
As shown in Table \ref{TAB1} and \ref{TAB2}, the ProbPlug outperforms several existing competitive confidence estimation methods on the same task in terms of both F1-score and AUPRC. Additionally, we evaluated our method on cross tasks without additional training by simply modifying prompts. The results demonstrate that our method outperforms current state-of-the-art approaches across most metrics. On the SST2 dataset, our method achieves a slightly lower F1-score than the CISC method. However, it exhibits significant advantages in other aspects: first, our method needs only one inference pass, with inherent efficiency advantages over CISC, which requires multiple repeated inferences. Furthermore, our method performs better in AUPRC, which implies that it could achieve higher performance by adjusting the classification threshold. Overall, the experiments demonstrate that our method achieves consistently improved performance on both in-task and cross-task datasets, with results averaged over multiple runs.

\begin{table}[t]
\centering
\caption{Comparison of the performance on the speech emotion recognition (Iemocap from the EmoBox benchmark \cite{ma2024emobox}) task with multimodal large model. UA means unweighted accuracy, and WA denotes the weighted accuracy. The performance of Non-LLM-based methods was cited from the corresponding papers.}
\label{TAB：emo}
\begin{tabular}{|l|l|c|c|c|}
\hline
\bfseries Type & \bfseries Methods & \bfseries UA & \bfseries WA & \bfseries F1-score \\
\hline
\multirow{3}{*}{Non-LLM-based} 
 & data2vec 2.0 large \cite{baevski2023efficient} & 57.30 & 56.23 & 56.70\\
 & Whisper large v3 \cite{radford2023robust} & 73.54 & 72.86 & 73.11 \\
 & Emotion2vec large \cite{ma2024emotion2vec}  & 70.70& 63.30 & / \\
\hline
\multirow{7}{*}{LLM-based} 
 & Qwen2-audio \cite{chu2024qwen2} & 64.33 &  60.37 & 61.61 \\
 & Verbalization \cite{tian2023just} & 66.45 & 63.24 & 62.56  \\
 & Logit \cite{yin2023large} & 66.42 & 64.35 & 61.26\\
 & Self-Consist \cite{wang2023selfconsistency} & 67.24 & 66.21 & 63.13 \\
 & CISC \cite{taubenfeld2025confidence} & 67.78 & 67.43 & 63.62 \\
 & SAPLMA \cite{azaria2023internal}  & 73.15 & 74.61 & 73.53 \\
 & \textbf{ProbPlug (Ours)} & \textbf{77.33} & \textbf{77.32} & \textbf{77.76}\\
\hline
\end{tabular}
\end{table}

\subsection{Multi-class Classification Performance}
Furthermore, to evaluate the performance of our method on multi-class classification tasks and multimodal large models, we trained our ProbPlug model using Qwen2-audio as the base model. As shown in Table \ref{TAB：emo}, our method exhibits significant advantages across all metrics compared to both existing non-LLM-based methods (e.g., Emotion2Vec) and other confidence estimation methods mentioned earlier. This indicates that our method can not only be generalized to multi-class classification tasks but also be applied to multimodal large models.

\begin{table}[t]
\centering
\caption{Calibration performance of SAPLMA and ProbPlug. }
\label{TAB:calibration}
\begin{tabular}{|l|c|c|c|c|}  
\hline
\multirow{2}{*}{\bfseries Calibration} & \multicolumn{2}{c|}{\bfseries ECE} & \multicolumn{2}{c|}{\bfseries Brier Score} \\  
\cline{2-5}  
 & \bfseries SAPLMA & \bfseries ProbPlug & \bfseries SAPLMA & \bfseries ProbPlug  \\  
\hline  
SMS-SPAM & 0.0194 & \textbf{0.0126} & 0.0240 & \textbf{0.0210} \\
SST2 & \textbf{0.1123} & 0.1251 & 0.0987 & \textbf{0.0706} \\
Toxic Comment Class & 0.0695 & \textbf{0.0654} & 0.0740 & \textbf{0.0739} \\
Civil comments & 0.1884 & \textbf{0.1580} & 0.2183 & \textbf{0.2092} \\
Amazon-polarity & 0.1015 & \textbf{0.0412} & 0.0563 & \textbf{0.0374} \\
\hline  
Average & 0.0982 & \textbf{0.0805} & 0.0943 & \textbf{0.0824} \\
\hline  
\end{tabular}
\end{table}

\begin{figure*}[t]  
  \centering 
  \includegraphics[width=1.0\textwidth]{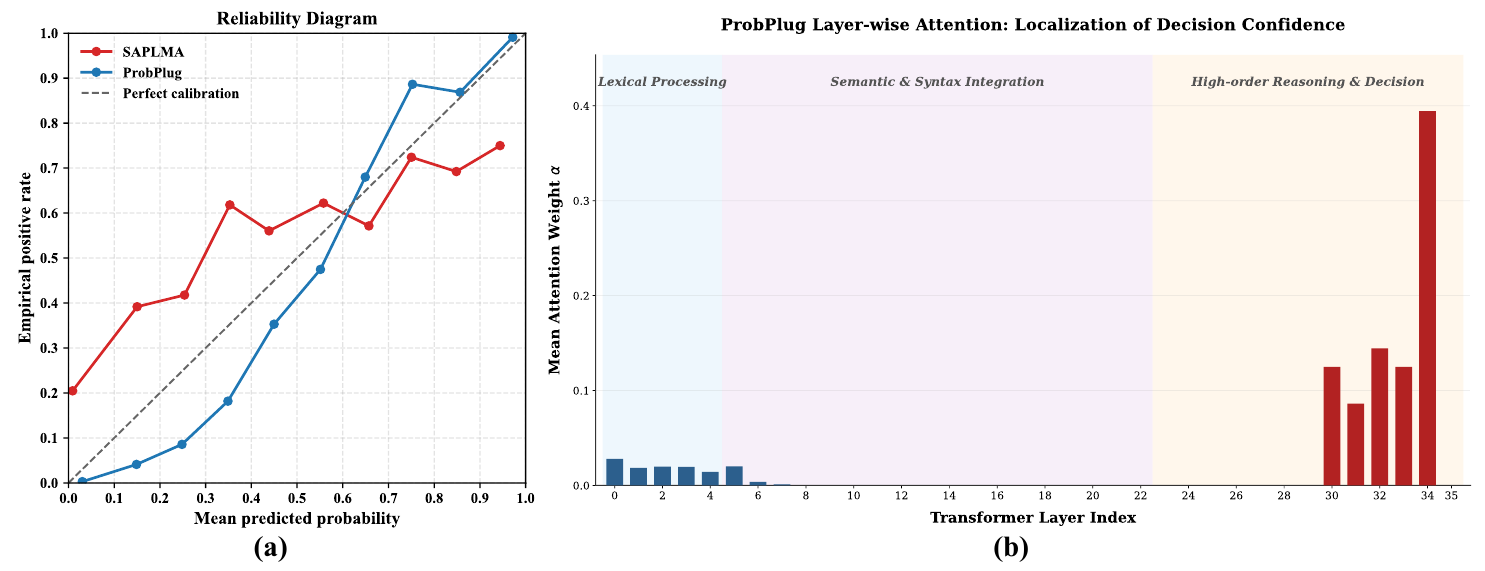} 
  \caption{(a)Reliability diagrams of confidence calibration for SAPLMA and ProbPlug on Amazon-polarity. (b)Visualization of the learned attention weights $\alpha$ across the 36-layer base model. The distribution demonstrates the concentration of decision certainty within the terminal high-order reasoning layers (L30–L35).} 
  \label{fig:reliability}
\end{figure*}

\subsection{Calibration Experiments}
To evaluate confidence reliability, we adopt Expected Calibration Error (ECE) and Brier Score as calibration metrics. 
ECE evaluates the consistency between predicted confidence and observed accuracy, 
whereas Brier Score quantifies the discrepancy between predicted probabilities and the corresponding ground-truth labels. 
As reported in Table \ref{TAB:calibration}, ProbPlug achieves lower average ECE (0.0805) and Brier Score (0.0824) than SAPLMA, indicating better calibration, with notable gains on SMS-SPAM and Amazon-polarity. Figure \ref{fig:reliability} (a) further shows that ProbPlug stays closer to the perfect calibration line across most bins, while SAPLMA exhibits clear overconfidence in high-confidence regions. These results demonstrate that ProbPlug provides more reliable confidence estimates.

\begin{table}[t]
\centering
\caption{Evaluation of in-task and cross-task performance under different ablation settings. TC means token compression module, while LIA means layer information aggregation module.}
\label{TAB6}
\begin{tabular}{|l|c|c|c|c|}  
\hline  
\multirow{2}{*}{\bfseries Configurations} & \multicolumn{2}{c|}{\bfseries In-task (SMS-SPAM)} & \multicolumn{2}{c|}{\bfseries Cross-task (SST2)} \\
\cline{2-5}  
 & \bfseries F1-score & \bfseries AUPRC & \bfseries F1-score & \bfseries AUPRC  \\
\hline  
w/o TC & 94.05 & 96.99 & 74.98 & 95.44 \\
w/o LIA & 92.26 & 95.16 & 86.53 & 96.31 \\
10 token & \bfseries 95.94 & \bfseries 97.84 & 86.84 & 94.95 \\
Ours & 94.78 & 97.56 & \bfseries 88.47 & \bfseries 97.42 \\
\hline  
\end{tabular}
\end{table}

\subsection{Ablation Studies}
\label{sec 4.5}

We further analyze the contribution of each model component and the effect of the token selection hyperparameter (number of tokens, denoted as $S$ in Section \ref{architecture}). Models with different configurations are trained on SMS-SPAM and evaluated on both SMS-SPAM and SST2. As shown in Table \ref{TAB6}, both the token compression module and the layer-wise information aggregation module improve performance. Using 10 tokens slightly improves in-task performance compared with (S=1), but noticeably reduces cross-task generalization. This is likely because preceding tokens contain task-specific semantic information, causing the confidence estimator to overfit to the source domain. In contrast, the final token mainly captures decision confidence with less task-specific noise. Therefore, using only the final token (S=1) provides better robustness across tasks.

\subsection{Layer-wise Attention Visualization and Interpretability}
To investigate where decision uncertainty is encoded across the 36 Transformer layers, we visualize the learned attention weights $\alpha$ under different inference conditions. As shown in Figure \ref{fig:reliability} (b), ProbPlug exhibits a clear bimodal pattern in high-confidence cases, with attention concentrated on the early lexical layers (L0–L5) and the final reasoning layers (L30–L35), while intermediate layers receive little weight.This suggests that confident predictions mainly rely on basic lexical cues and high-level reasoning, whereas intermediate semantic representations contribute limited information and may introduce task-specific noise. By suppressing these less informative layers, ProbPlug maintains a focused attention distribution and improves confidence robustness. The results indicate that the learned layer-wise attention effectively reflects the evolution of decision uncertainty across the Transformer hierarchy.

\section{Conclusion}
In this work, we introduce ProbPlug, a plug-in framework that estimates prediction confidence for LLM-based binary classification from hidden token representations. By operating on internal representations of the frozen base model without modifying its original architecture, ProbPlug addresses the problem of uncalibrated confidence in LLM outputs.
Extensive experiments on multiple classification tasks, using both standard large language models (e.g., Qwen3) and multimodal large models (e.g., Qwen2-Audio), show that ProbPlug achieves three main benefits. First, it provides more reliable confidence estimates, improving the trustworthiness of LLM predictions. Second, it brings consistent gains in downstream classification performance with only limited additional inference overhead. Third, it demonstrates promising cross-task generalization, reducing the need for extensive retraining when adapting to new tasks.
Overall, the experimental results demonstrate that ProbPlug offers an effective approach to confidence estimation for LLM-based classification. This study further highlights its potential to support more reliable deployment of LLMs in high-stakes scenarios that require trustworthy confidence assessment.

%
%
%
%

\end{document}